\documentclass{article}

\PassOptionsToPackage{numbers, compress}{natbib}
\usepackage[preprint]{neurips_2023}

\usepackage[utf8]{inputenc} % allow utf-8 input
\usepackage[T1]{fontenc}    % use 8-bit T1 fonts
\usepackage{hyperref}       % hyperlinks
\usepackage{url}            % simple URL typesetting
\usepackage{booktabs}       % professional-quality tables
\usepackage{amsfonts}       % blackboard math symbols
\usepackage{nicefrac}       % compact symbols for 1/2, etc.
\usepackage{microtype}      % microtypography
\usepackage{xcolor}         % colors

\usepackage{graphicx}
\usepackage{adjustbox,multirow}
\usepackage{tcolorbox}

\newcommand{\oursdataset}{APIBench}
\newcommand{\oursmethod}{Gorilla}
\newcommand{\gorilla}{Gorilla}

\title{Gorilla: Large Language Model Connected with Massive APIs}

\author{%
  Shishir G. Patil$^{1}$\thanks{Equal contribution.} \quad Tianjun Zhang$^{1,*}$ \quad Xin Wang$^{2}$ \quad  Joseph E. Gonzalez$^{1}$ \\
$^1$UC Berkeley \qquad $^2$Microsoft Research \\
sgp@berkeley.edu
}

\begin{document}

\maketitle

\begin{abstract}
  Large Language Models (LLMs) have seen an impressive wave of advances recently, with models now excelling in a variety of tasks, such as mathematical reasoning and program synthesis. However, their potential to effectively use tools via API calls remains unfulfilled. This is a challenging task even for today's state-of-the-art LLMs such as GPT-4, largely due to their inability to generate accurate input arguments and their tendency to hallucinate the wrong usage of an API call. We release Gorilla, a finetuned LLaMA-based model that surpasses the performance of GPT-4 on writing API calls. When combined with a document retriever, Gorilla demonstrates a strong capability to adapt to test-time document changes, enabling flexible user updates or version changes. It also substantially mitigates the issue of hallucination, commonly encountered when prompting LLMs directly. To evaluate the model's ability, we introduce APIBench, a comprehensive dataset consisting of HuggingFace, TorchHub, and TensorHub APIs. The successful integration of the retrieval system with Gorilla demonstrates the potential for LLMs to use tools more accurately, keep up with frequently updated documentation, and consequently increase the reliability and applicability of their outputs.
  Gorilla's code, model, data, and demo are available at \url{https://gorilla.cs.berkeley.edu}
\end{abstract}

\section{Introduction}
\label{sec:intro}

Recent advances in large language models (LLMs)~\cite{chowdhery2022palm, brown2020language, scao2022bloom, bubeck2023sparks, openai2023gpt4, ChatGPT} have enabled significant new capabilities including natural dialogue, mathematical reasoning, and program synthesis. 
However, despite these advances, LLMs are still fundamentally limited by the information they can store in a fixed set of weights and the things they can compute using a static computation graph and limited context. 
Furthermore, as the world changes, LLMs require retraining to update their knowledge and reasoning capabilities. 

By empowering LLMs to use tools~\cite{schick2023toolformer}, we can grant access to vastly larger and changing knowledge bases and accomplish complex computational tasks.
By providing access to search technologies and databases, \cite{nakano2021webgpt, thoppilan2022lamda, shuster2022blenderbot} demonstrated that we can augment LLMs to address a significantly larger and more dynamic knowledge space. 
Similarly, by providing access to computational tools, \cite{thoppilan2022lamda, andor2019giving} demonstrated that LLMs can accomplish complex computational tasks. 
Consequently, leading LLM providers\cite{openai2023gpt4}, have started to integrate plugins to allow LLMs to invoke external tools through APIs.

This transition from a small set of hand-coded tools, to the ability to invoke a vast space of changing cloud APIs could transform LLMs into the primary interface to computing infrastructure and the web. 
Tasks ranging from booking an entire vacation to hosting a conference, could become as simple as talking to an LLM that has access to the flight, car rental, hotel, catering, and entertainment web APIs. 
However, much of the prior work~\cite{shen2023hugginggpt, liang2023taskmatrix} integrating tools into LLMs considered a small well documented set of APIs that can be easily injected into the prompt.

Supporting a web scale collection of potentially millions of changing APIs requires rethinking our approach to how we integrate tools.
It is not longer possible to describe the full set of APIs in a single context. 
Many of the APIs will have overlapping functionality with nuanced limitations and constraints. 
Simply evaluating LLMs in this new setting requires new benchmarks.

\begin{figure}[t]
    \includegraphics[width=\linewidth]{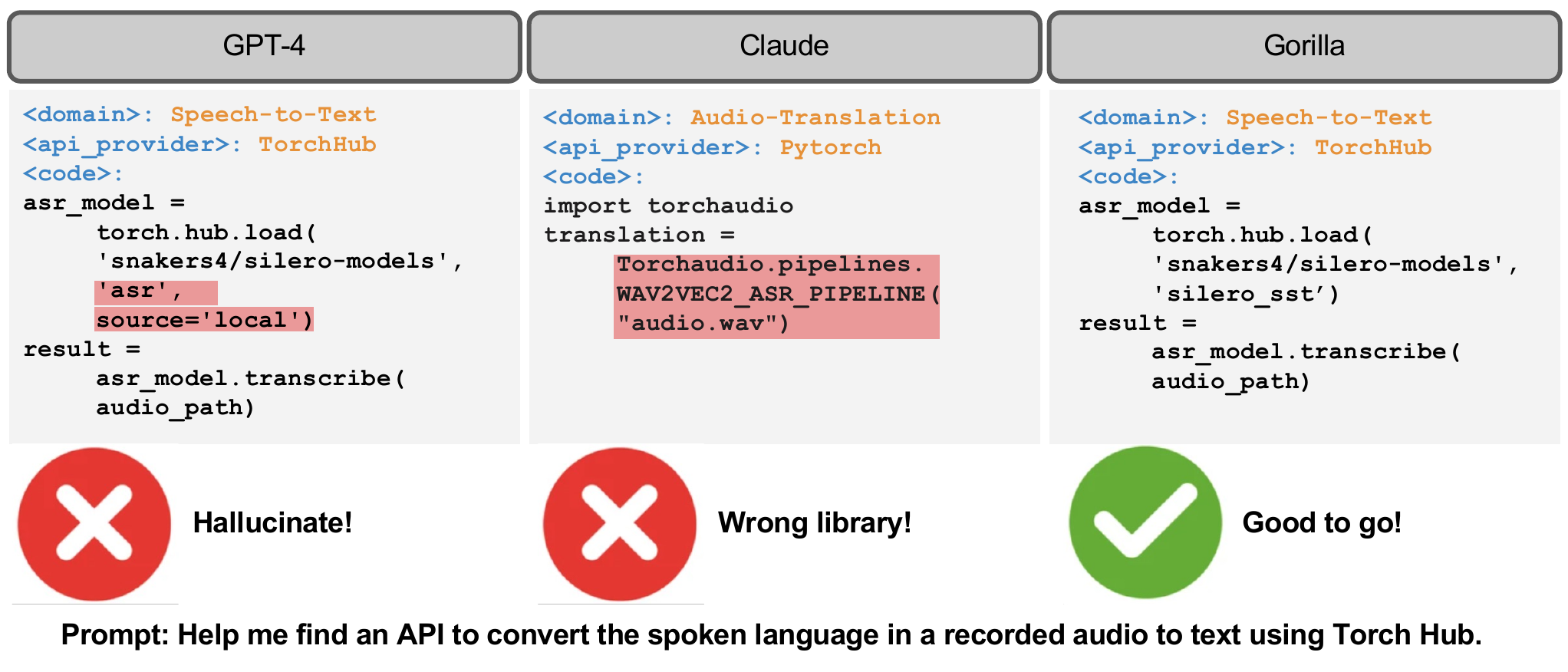}
\caption{\footnotesize \textbf{Examples of API calls}. Example API calls generated by GPT-4~\cite{openai2023gpt4}, Claude~\cite{Claude}, and \oursmethod{} for the given prompt. In this example, GPT-4 presents a model that doesn't exist, and Claude picks an incorrect library. In contrast, our \oursmethod{} model can identify the task correctly and suggest a fully-qualified API call.}
\label{fig:examplecode}
\end{figure}

In this paper, we explore the use of self-instruct fine-tuning and retrieval to enable LLMs to accurately select from a large, overlapping, and changing set tools expressed using their APIs and API documentation.
We construct, \oursdataset{}, a large corpus of APIs with complex and often overlapping functionality by scraping ML APIs (models) from public model hubs.  
We choose three major model hubs for dataset construction: TorchHub, TensorHub and HuggingFace. We exhaustively include every API call in TorchHub (94 API calls) and TensorHub (696 API calls); For HuggingFace, since the models come in a large number and lots of the models don't have a specification, we choose the most downloaded 20 models per task category (in a total of 925). We also generate 10 synthetic user question prompts per API using Self-Instruct~\cite{wang2022self}. Thus, each entry in the dataset becomes an instruction reference API pair. We adopt a common AST sub-tree matching technique to evaluate the functional correctness of the generated API. We first parse the generated code into an AST tree, then find a sub-tree whose root node is the API call that we care about (e.g., \texttt{torch.hub.load}) and use it to index our dataset. We check the functional correctness and hallucination problem for the LLMs, reporting the corresponding accuracy. 

We then finetune \oursmethod{}, a LLaMA-7B-based model with document retrieval using our dataset. We find that \oursmethod{} significantly outperforms GPT-4 in terms of API functionality accuracy as well as reducing hallucination errors. We show an example output in Fig.~\ref{fig:examplecode}.  Further, our retrieval-aware training of \gorilla{} enables the model to adapt to changes in the API documentation. Finally, we demonstrate Gorilla's ability to understand and reason about constraints.

\begin{figure}[t]
    \includegraphics[width=\linewidth]{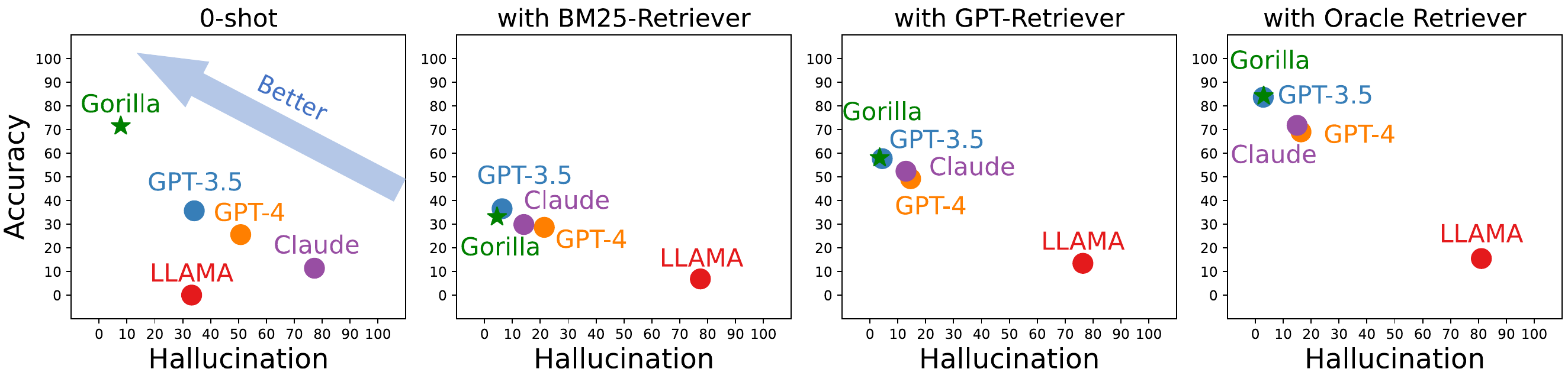}
\caption{\footnotesize \textbf{Accuracy (vs) hallucination} in four settings, that is,  \emph{zero-shot} (i.e., without any retriever), and \emph{with retrievers}. \texttt{BM25} and \texttt{GPT} are commonly used retrievers and the \texttt{oracle} retriever returns relevant documents at 100\%, indicating an upper bound. Higher in the graph (higher accuracy) and to the left is better (lower hallucination). Across the entire dataset, our model, \oursmethod{}, improves accuracy while reducing hallucination.}
\label{fig:acc-hallu}
\end{figure}

\section{Related Work}
\label{sec:related}

\paragraph{Large Language Models}
Recent strides in the field of LLMs have renovated many downstream domains~\cite{chowdhery2022palm, touvron2023llama, zhang2022opt, zeng2022glm}, not only in traditional natural language processing tasks but also in program synthesis. Many of these advances are achieved by augmenting pre-trained LLMs by prompting~\cite{wei2022chain, gao2022pal} and instruction fine-tuning~\cite{chung2022scaling, sanh2021multitask, wang2022super, iyer2022opt}. Recent open-sourced models like
LLaMa~\cite{touvron2023llama}, Alpaca~\cite{alpaca}, and Vicuna~\cite{vicuna} have furthered the understanding of LLMs and facilitated their experimentation. While our approach, \oursmethod{}, incorporates techniques akin to those mentioned, its primary emphasis is on enhancing the LLMs' ability to utilize millions of tools, as opposed to refining their conversational skills. Additionally, we pioneer the study of fine-tuning a base model by supplementing it with information retrieval - a first, to the best of our knowledge.

\paragraph{Tool Usage}
The discussion of tool usage within LLMs has seen an upsurge, with models like Toolformer taking the lead~\cite{schick2023toolformer, komeili2021internet, lazaridou2022internet, nakano2021webgpt}. Tools often incorporated include web-browsing~\cite{schick2020exploiting}, calculators~\cite{cobbe2021training, thoppilan2022lamda}, translation systems~\cite{thoppilan2022lamda}, and Python interpreters~\cite{gao2022pal}. 
While these efforts can be seen as preliminary explorations of marrying LLMs with tool usage, they generally focus on specific tools. Our paper, in contrast, aims to explore a vast array of tools (i.e., API calls) in an open-ended fashion, potentially covering a wide range of applications.

With the recent launch of Toolformer~\cite{schick2023toolformer} and GPT-4~\cite{openai2023gpt4}, the importance of API calls has been highlighted, encouraging many works in employing API calls as tooling
~\cite{shen2023hugginggpt, liang2023taskmatrix}. Moreover, the application of API calls in robotics has been explored to some extent~\cite{vemprala2023chatgpt, ahn2022can}. However, these works primarily aim at showcasing the potential of ``prompting'' LLMs rather than establishing a systematic method for evaluation and training (including fine-tuning). Our work, on the other hand, concentrates on systematic evaluation and building a pipeline for future use.

\paragraph{LLMs for Program Synthesis}
Harnessing LLMs for program synthesis has historically been a challenging task~\cite{li2022competition, chen2021evaluating, xu2022systematic, jain2022jigsaw, devlin2017robustfill, lachaux2020unsupervised}. Researchers have proposed an array of strategies to prompt LLMs to perform better in coding tasks, including in-context learning~\cite{wei2022chain, kojima2022large, chen2021evaluating}, task decomposition~\cite{kim2023language, yao2022react}, and self-debugging~\cite{chen2023teaching, shinn2023reflexion}. Besides prompting, there have also been efforts to pretrain language models specifically for code generation~\cite{nijkamp2022codegen, li2023starcoder, nijkamp2023codegen2}.

However, these strategies focus on prompting large language models or pre-training them for general program synthesis. In our research, in contrast, we focus on a much restricted domain: the synthesis of linear programs using API calls. General program synthesis, not only is complex, but is also hard to verify and evaluate. API calls, on the other hand, function more like tool usage. This allows the LLM to significantly expand its capabilities without grappling with low-level implementation details.
\section{Methodology}
\label{sec:method}

\begin{figure}[t]
    \includegraphics[width=\linewidth]{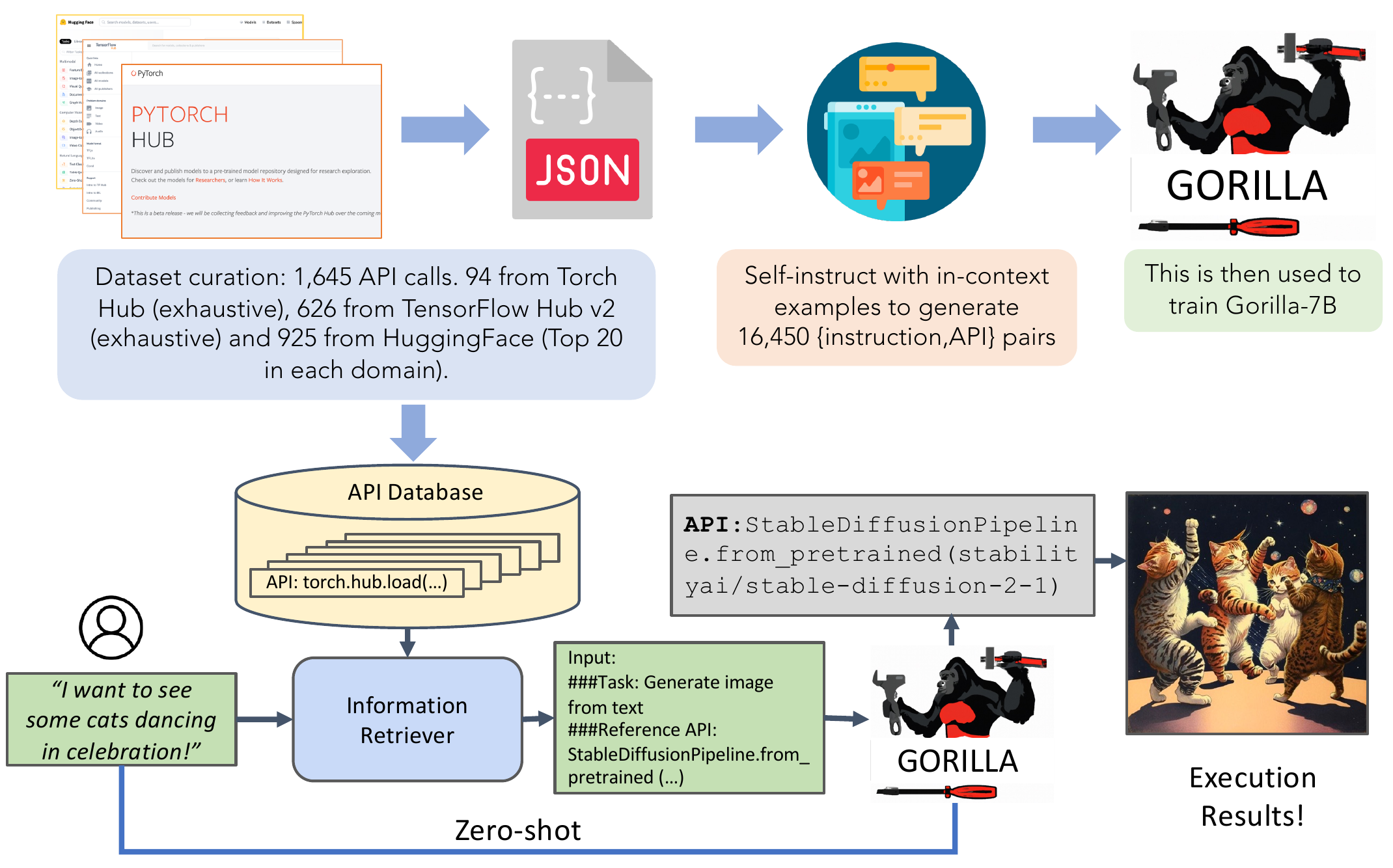}
\caption{\footnotesize \textbf{Gorilla: A system for enabling LLMs to interact with APIs.} The upper half represents the training procedure as described in Sec~\ref{sec:method}. This is the most exhaustive API data-set for ML to the best of our knowledge. During inference (lower half), \gorilla{} supports two modes - with retrieval, and zero-shot. In this example, it is able to suggest the right API call for generating the image from the user's natural language query.}
\label{fig:gorilla}
\end{figure}

In this section, we describe \oursdataset{}, a comprehensive benchmark constructed from TorchHub, TensorHub, and HuggingFace API Model Cards. We begin by outlining the process of collecting the API dataset and how we generated instruction-answer pairs. We then introduce \oursmethod{}, a novel training paradigm with a  information\---retriever incorporated into the training and inference pipelines. Finally, we present our AST tree matching evaluation metric.

\subsection{Dataset Collection}
To collect the dataset, we meticulously recorded all online model cards for HuggingFace's ``The Model Hub'', PyTorch Hub, and TensorFlow Hub Models. Throughout the rest of the paper, we call these HuggingFace, Torch Hub, and TensorFlow Hub respectively for brevity. 

\paragraph{API Documentation} The HuggingFace platform hosts and servers about 203,681 models. However, many of them have poor documentation, lack dependencies, have no information in their model card, etc. To filter these out, we pick the top 20 models from each domain. 
We consider 7 domains in multimodal data, 8 in CV, 12 in NLP, 5 in Audio, 2 in tabular data, and 2 in  reinforcement learning. Post filtering, we got a total of 925 models from HuggingFace. TensorFlow Hub is versioned into v1 and v2. The latest version (v2) has 801 models in total, and we process all of them. Post filtering out models, whose mode cards had little to no information, we are left with 626 models. Similar to TensorFlow Hub, we get 95 models from Torch Hub. 
We then converted the model cards for each of these 1,645 API calls into a json object with the following fields: \{domain, framework, functionality, api\_name, api\_call, api\_arguments, environment\_requirements, example\_code, performance, and description.\}. We provide more information in the Appendix. These fields were chose to generalize beyond the API calls within ML domain, to other domains, includin RESTful API calls.

\paragraph{Instruction Generation}
Guided by the self-instruct paradigm~\cite{wang2022self}, we employed GPT-4 to generate synthetic instruction data. 
We provided three in-context examples, along with a reference API documentation, and tasked the model with generating real-world use cases that call upon the API. We specifically instructed the model to refrain from using any API names or hints when creating instructions. We constructed six examples (Instruction-API pairs) for each of the three model hubs. These 18 points, were the only hand-generated or modified data. For each of our 1,645 API datapoints, we sample 3 of 6 corresponding instruction examples to generate a total of 10 instruction-api pairs as demonstrated in Figure~\ref{fig:gorilla}. We would like to highlight that we only need to employ GPT-4 to generate the instructions and this can be swapped  with open-source alternatives such as LLaMA, Alpaca, etc. 

\subsection{\gorilla{}}
Our model \oursmethod{}, is retrieve-aware finetuned LLaMA-7B model, specifically for API calls. As shown in Fig~\ref{fig:gorilla}, we employ self-instruct to generate \{instruction, API\} pairs. To fine-tune LLaMA, we convert this to a user-agent chat-style conversation, where each data-point is a conversation with one round each for the user and the agent. We then perform standard instruction finetuning on the base LLaMA-7B model. For our experiments, we train \gorilla{} with and without the retriever. 

\paragraph{API Call with Constraints} API calls often come with inherent constraints. These constraints necessitate that the LLM not only comprehend the functionality of the API call but also categorize the calls according to different constraint parameters. 
This requirement introduces an additional layer of complexity to the process, demanding a more nuanced understanding from the LLM.
Specifically, for machine learning API calls, two common sets of constraints are: parameter size and a lower bound on accuracy. Consider, for instance, the following prompt: \texttt{``Invoke an image classification model that uses less than 10M parameters, but maintains an ImageNet accuracy of at least 70\%''}. Such a prompt presents a substantial challenge for the LLM to accurately interpret and respond to. Not only must the LLM understand the user's functional description, but it also needs to reason about the various constraints embedded within the request. This challenge underlines the intricate demands placed on LLMs in real-world API calls. It is not sufficient for the model to merely comprehend the basic functionality of an API call; it must also be capable of navigating the complex landscape of constraints that accompany such calls. These observations necessitate the need to fine-tune an LLM for APIs.

\paragraph{Retriever-Aware training} For training with retriever, the instruction-tuned dataset, also has an additional \texttt{"Use this API documentation for reference: <retrieved\_API\_doc\_JSON>"} appended to the user prompt. Through this, we aim to teach the LLM to parse the second half of the question to answer the first half. We demonstrate that this a) makes the LLM adapt to test-time changes in API documentation, and b) improves performance  from in-context learning, and finally c) show that it reduces hallucination error. 

Surprisingly, we find that augmenting a LLM with retrieval, does not always lead to improved performance, and can at-times hurt performance. We share more insights along with details in Sec~\ref{sec:eval}.

\paragraph{\oursmethod{} Inference} During Inference, the user provides the prompt in natural language (Fig:~\ref{fig:gorilla}). This can be for a simple task (e.g, \emph{``I would like to identify the objects in an image''}), or they can specify a vague goal, (.e.g, \emph{``I am going to the zoo, and would like to track animals''}). \gorilla{}, similar to training, can be used for inference in two modes: zero-shot and with retrieval. In zero-shot, this prompt (with NO further prompt tuning) is fed to the \gorilla{} LLM model when then returns the API call that will help in accomplishing the task and/or goal. In retrieval mode, the retriever (either of BM25 or GPT-Index) first retrieves the most up-to-date API documentation stored in the API Database. This is then concatenated to the user prompt along with the message \texttt{Use this API documentation for reference:} before feeding it to \gorilla{}.  The output of \gorilla{} is an API to be invoked. Besides the concatenation as described, we do \emph{NO} further prompt tuning in our system. While we do have a system to execute these APIs, that is not a focus of this paper.

\subsection{Verifying APIs}

Inductive program synthesis, where a program is synthesized to satisfy test cases, has found success in several avenues~\cite{autopandas, flashfill}. 
However, test cases fall short when evaluating API calls, as it is often hard to verify the semantic correctness of the code. For example, consider the task of classifying an image. 
There are over 40 different models that can be used for the task. Even if we were to narrow down to a single family of Densenet, there are four different configurations possible. Hence, there exist multiple correct answers and it is hard to tell if the API being used is functionally equivalent to the reference API by unit tests. Thus, to evaluate the performance of our model, we compare their functional equivalence using the dataset we collected. To trace which API in the dataset is the LLM calling, we adopt the AST tree-matching strategy. Since we only consider one API call in this paper, checking if the AST of the candidate API call is a sub-tree of the reference API call reveals which API is being used in the dataset.

Identifying and even defining hallucinations can be challenging. 
We use the AST matching process to directly identify the hallucinations. 
We define a hallucination as an API call that is not a sub-tree of any API in the database -- invoking an entirely imagined tool.
This form of hallucination is distinct from invoking an API incorrectly which we instead define as an error.

\begin{figure}[t]
    \includegraphics[width=\linewidth]{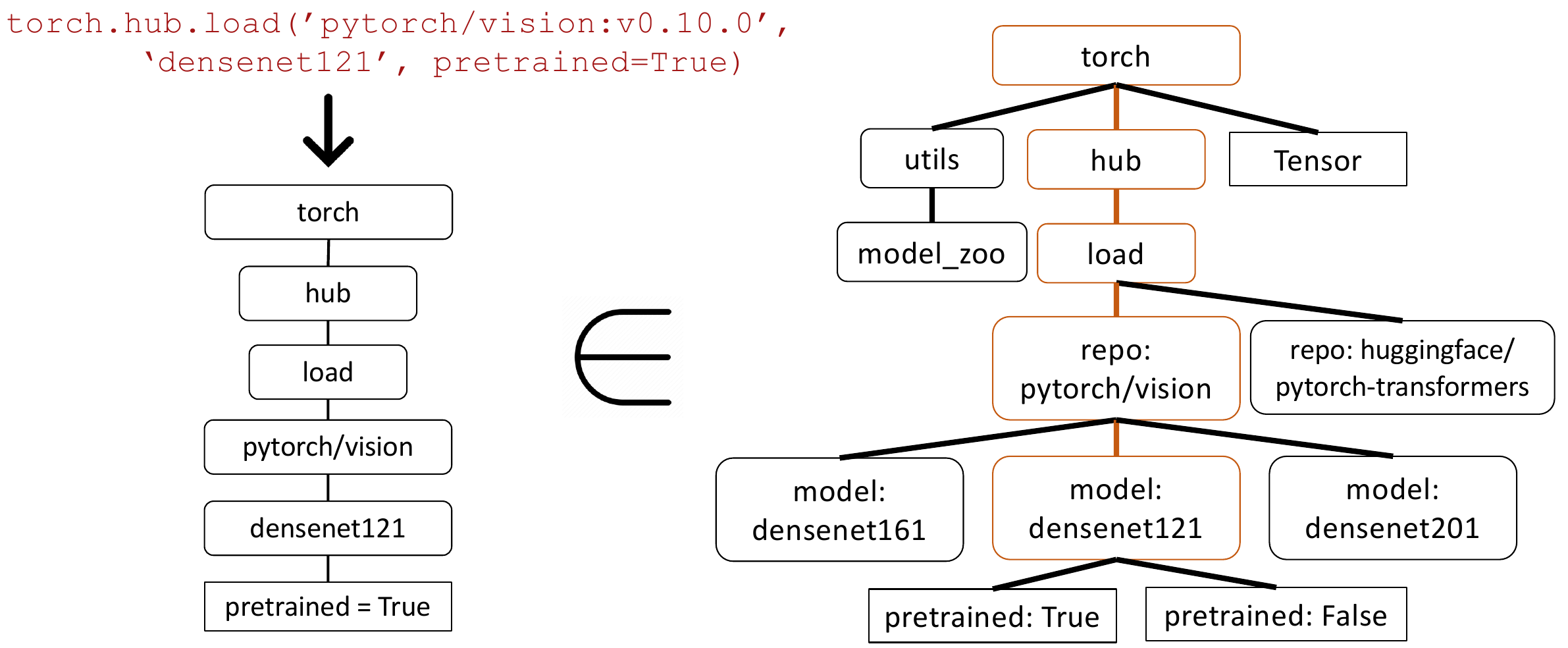}
\caption{\footnotesize \textbf{AST Sub-Tree Matching to evaluate API calls.} On the left is an API call returned by \gorilla{}. We first build the associated API tree. We then compare this to our dataset, to see if the API dataset has a sub\-tree match. In the above example, the matching sub\-tree is highlighted in brown, signifying that the API call is indeed correct. \texttt{Pretrained=True} is an optional argument.}
\label{fig:ast}
\end{figure}

\paragraph{AST Sub-Tree Matching} We perform AST sub-tree matching to identify which API in our dataset is the LLM calling. Since each API call can have many arguments, we need to match on each of these arguments. Further, since, Python allows for default arguments, for each API, we define which arguments to match in our database. For example, we check \texttt{repo\_or\_dir} and \texttt{model} arguments in our function call. In this way, we can easily check if the argument matches the reference API or not. Please refer to Fig.~\ref{fig:ast} for more details. In this example, \gorilla{} returns a torch API call. We first build the tree, and verify that it matches a sub\-tree in our dataset along nodes \texttt{torch.hub.load}, \texttt{pytorch/vision}, and \texttt{densenet121}. But, we don't check for match along leaf node \texttt{pretrained = True} since that is an optional python argument.

\section{Evaluation}
\label{sec:eval}
We carried out an array of experiments on our collected dataset, benchmarking our model \gorilla{} with other models, and exploring how different retrieval methods may impact the performance of the model in making API calls. We then demonstrate that \gorilla{} can easily adapt to test-time changes in API documentation. In addition, we assess Gorilla's ability to reason about API calls under constraints. Lastly, we examined how integrating different retrieval methods during training influences the model's final performance.

\paragraph{Baselines} Primarily, we compare \oursmethod{} with state-of-the-art language models in a zero-shot setting. The models under consideration include: GPT-4 by OpenAI, we use the \texttt{gpt-4-0314} checkpoint; GPT-3.5-turbo with the \texttt{gpt-3.5-turbo-0301} checkpoint, both of which are RLHF-tuned model specifically designed for conversation; Claude with \texttt{claude-v1} checkpoint, a language model by Anthropic, renowned for its lengthy context capabilities; LLaMA-7B, a large language model by Meta and the finest open-source model to date.

\paragraph{Retrievers} The term \emph{Zero-shot} (abbreviated as 0-shot in tables) refers to scenarios where no retriever is used. The sole input to the model is the user's natural language prompt. For \texttt{BM25}, we consider each API as a separate document. During retrieval, we use the user's query to search the index and fetch the most relevant (top-1) API. 
This API is concatenated with the user's prompt to query the LLMs. Similarly, GPT-Index refers to the retrieval model \texttt{text-davinci-003} from OpenAI. Like BM25, each API call is indexed as an individual document, and the most relevant document, given a user query, is retrieved and appended to the user prompt. Lastly, we include an Oracle retriever, which serves two purposes: first, to identify the potential for performance improvement through more efficient retrievers, and second, to assist users who know which API to use but may need to help invoking it. 
In all cases, when a retriever is used, it is appended to the user's prompt as follows: \texttt{<user\_prompt>} \texttt{Use this API documentation for reference:} \texttt{<retrieved\_API\_doc\_JSON>}. 
The dataset for these evaluations is detailed in Sec~\ref{sec:method}. We emphasize that we have maintained a holdout test set on which we report our findings. The holdout test set was created by dividing the self-instruct dataset's {instruction, API} pairs into training and testing sets.

\subsection{AST Accuracy on API call}

\begin{figure}[t]
    \includegraphics[width=\linewidth]{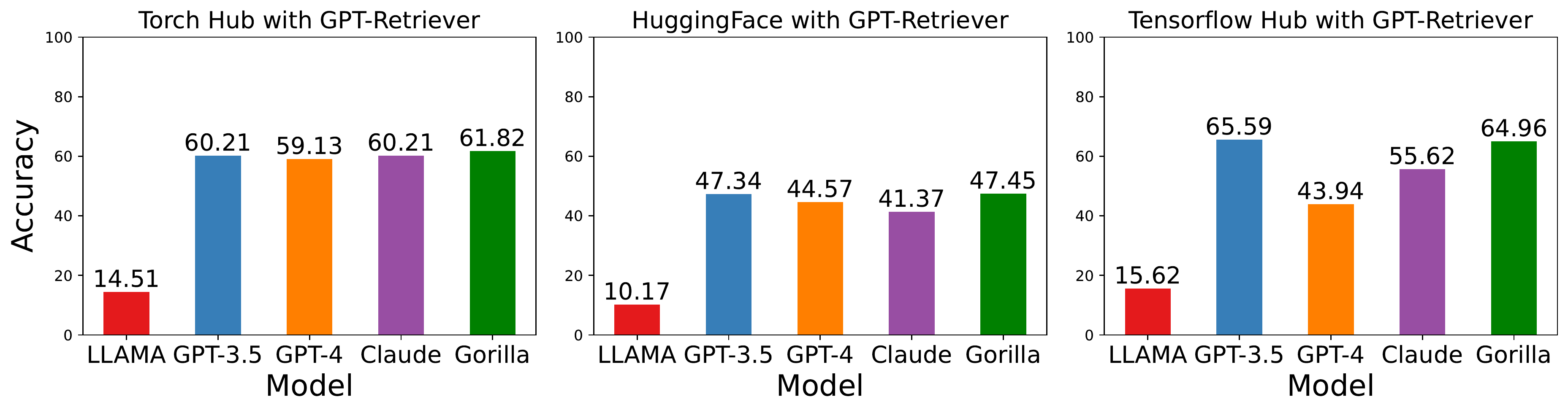}
\caption{\footnotesize \textbf{Accuracy with GPT-retriever.} 
\oursmethod{} outperforms on Torch Hub and Hugging-Face while matching performance on Tensorflow Hub for all existing SoTA LLMs - closed source, and open source. }
\label{fig:images}
\end{figure}
We first demonstrate the results for the AST accuracy for different models. We present the results in Tab.~\ref{tab:llm_eval}. We test each model for different retriever settings defined above. We report the overall accuracy, the error by hallucination and the error by selecting wrong API call. Note that for TorchHub and TensorHub, we evaluate all the models using AST tree accuracy score. However, for HuggingFace, since the dataset is not exhaustive, for all the models except \oursmethod{}, we only check if they can provide the correct domain names. So this problem reduces to picking one of the multiple choices.

\paragraph{Finetuning without Retrieval} In Tab.~\ref{tab:llm_eval} we show that lightly fine-tuned \oursmethod{} gets the state-of-the-art performance zero-shot over all the models, 20.43\% better than GPT-4 and 10.75\% better than ChatGPT. When compared to other open-source models LLAMA, the improvement is as big as 83\%. his suggests quantitatively, that finetuning is better than retrieval, at-least in our scope. 

In addition, we found that finetuning without retriever and putting ground truth retriever in evaluation time rarely helps the performance: 0.88\% worse in TensorHub and 0.97\% better in HuggingFace. If we put BM25 or GPT-Index as retriever, results will be significantly dropped: 21.50\% in Torch Hub and 47.57\% in HuggingFace. The result illustrates that adding a non-optimal retriever at test time will sometime misguide the model and result in more errors. We will discuss an interesting ablation on how finetuning with the retriever will help the performance in the next paragraph.

\begin{table*}[h]
    \caption{\small \textbf{Evaluating LLMs on Torch Hub, HuggingFace, and Tensorflow Hub APIs}}
    \label{tab:llm_eval}
     \setlength{\tabcolsep}{10pt} % Default value: 6pt
    \begin{adjustbox}{width=\textwidth}
    \begin{tabular}{l c c c | c c c | c c c}
    \toprule
    
    LLM (retriever) & \multicolumn{3}{c}{TorchHub} & \multicolumn{3}{c}{HuggingFace} & \multicolumn{3}{c}{TensorFlow Hub} \\
    & overall $\uparrow$ & hallu $\downarrow$ & err $\downarrow$ & overall $\uparrow$ & hallu $\downarrow$ & err $\downarrow$ & overall $\uparrow$ & hallu $\downarrow$ & err $\downarrow$ \\
    \midrule
    LLAMA (0-shot) & 0 & 100 & 0 & 0.00 & 97.57 & 2.43 & 0 & 100 & 0 \\
    GPT-3.5 (0-shot) & 48.38 & 18.81 & 32.79 & 16.81 & 35.73 & 47.46 & 41.75 & 47.88 & 10.36 \\
    GPT-4 (0-shot) & 38.70 & 36.55 & 24.7 & 19.80 & 37.16 & 43.03 & 18.20 & 78.65 & 3.13 \\
    Claude (0-shot) & 18.81 & 65.59 & 15.59 & 6.19 & 77.65 & 16.15 & 9.19 & 88.46 & 2.33 \\
    \gorilla{} (0-shot) & \textbf{59.13} & \textbf{6.98} & 33.87 & \textbf{71.68} & \textbf{10.95} & 17.36 & \textbf{83.79} & \textbf{5.40} & 10.80 \\
    \midrule
    LLAMA (BM-25) & 8.60 & 76.88 & 14.51 & 3.00 & 77.99 & 19.02 & 8.90 & 77.37 & 13.72 \\
    GPT-3.5 (BM-25) & 38.17 & 6.98 & 54.83 & \textbf{17.26} & 8.30 & 74.44 & \textbf{54.16} & 3.64 & 42.18 \\
    GPT-4 (BM-25) & 35.48 & 11.29 & 53.22 & 16.48 & 15.93 & 67.59 & 34.01 & 37.08 & 28.90 \\
    Claude (BM-25) & 39.78 & 5.37 & 54.83 & 14.60 & 15.82 & 69.58 & 35.18 & 21.16 & 43.64 \\
    \gorilla{} (BM-25) & \textbf{40.32} & \textbf{4.30} & 55.37 & 17.03 & \textbf{6.42} & 76.55 & 41.89 & \textbf{2.77} & 55.32 \\
    \midrule
    LLAMA (GPT-Index) & 14.51 & 75.8 & 9.67 & 10.18 & 75.66 & 14.20 & 15.62 & 77.66 & 6.71 \\
    GPT-3.5 (GPT-Index) & 60.21 & 1.61 & 38.17 & 29.08 & 7.85 & 44.80 & \textbf{65.59} & 3.79 & 30.50 \\
    GPT-4 (GPT-Index) & 59.13 & 1.07 & 39.78 & 44.58 & 11.18 & 44.25 & 43.94 & 31.53 & 24.52 \\
    Claude (GPT-Index) & 60.21 & 3.76 & 36.02 & 41.37 & 18.81 & 39.82 & 55.62 & 16.20 & 28.17 \\
    \gorilla{} (GPT-Index) & \textbf{61.82} & \textbf{0} & 38.17 & \textbf{47.46} & \textbf{8.19} & 44.36 & 64.96 & \textbf{2.33} & 32.70 \\
    \midrule
    LLAMA (Oracle) & 16.12 & 79.03 & 4.83 & 17.70 & 77.10 & 5.20 & 12.55 & 87.00 & 0.43 \\
    GPT-3.5 (Oracle) & 66.31 & 1.60 & 32.08 & 89.71 & 6.64 & 3.65 & \textbf{95.03} & \textbf{0.29} & 4.67 \\
    GPT-4 (Oracle) & 66.12 & 0.53 & 33.33 & 85.07 & 10.62 & 4.31 & 55.91 & 37.95 & 6.13 \\
    Claude (Oracle) & 63.44 & 3.76 & 32.79 & 77.21 & 19.58 & 3.21 & 74.74 & 21.60 & 3.64 \\
    \gorilla{} (Oracle) & \textbf{67.20} & \textbf{0} & 32.79 & \textbf{91.26} & \textbf{7.08} & 1.66 & 94.16 & 1.89 & 3.94 \\
    \bottomrule
    \end{tabular}
    \end{adjustbox}
\end{table*}

\paragraph{Finetuning with Retrieval} We now discuss an interesting experiment on how finetuning language with retriever incorporated is helping the performance. The settings for this experiment are finetuning the base LLAMA with the prompt (instruction generated), reference API document (from golden-truth oracle), and the example output generated by GPT-4. In Tab.~\ref{tab:retrieve}, we can see that incorporating ground truth retriever in the finetuning pipeline achieves significantly better results 12.37\% better than training without retriever in Torch Hub and 23.46\% better in HuggingFace. However, we found that at evaluation time, current retrievers still have a big gap between the ground truth retriever: using GPT-Index at evaluation results in 29.20\% accuracy degradation, and using BM25 results in a 52.27\% accuracy degradation. Nevertheless, we can still conclude that with a better retriever, finetuning with retriever is still a better method to adopt while in another scenario, when a good retriever is not available, zero-shot finetuning might be the preferred choice.

\begin{table*}[h]
    \caption{\small \textbf{Comparison of retrieval techniques}
    }
    \label{tab:retrieve}
     \setlength{\tabcolsep}{10pt} % Default value: 6pt
    \begin{adjustbox}{width=\textwidth}
    \begin{tabular}{l cccc | cccc}
    \toprule
    & \multicolumn{4}{c}{\gorilla{} without Retriever} & \multicolumn{4}{c}{\oursmethod{} with Oracle retriever} \\ 
     % \cmidrule(lr){4-23}
     \midrule
    & zero-shot & BM25 & GPT-Index & Oracle & zero-shot & BM25 & GPT-Index & Oracle  \\
    \midrule
    Torch Hub (overall) $\uparrow$ & 59.13 & 37.63 & 60.21 & 54.83 & 0 & 40.32 & 61.82 & 67.20  \\
    HuggingFace (overall) $\uparrow$ & 71.68 & 11.28 & 28.10 & 45.58 & 0 & 17.04 & 47.46 & 91.26  \\
    TensorHub (overall)  $\uparrow$ & 83.79 & 34.30 & 52.40 & 82.91 & 0 & 41.89 & 64.96 & 94.16 \\
    \midrule
    Torch Hub (Hallu) $\downarrow$ & 6.98 & 11.29 & 4.30 & 15.59 & 100 & 4.30 & 0 & 0  \\
    HuggingFace (Hallu) $\downarrow$ & 10.95 & 46.46 & 41.48 & 52.77 & 99.67 & 6.42 & 8.19 & 7.08  \\
    TensorHub (Hallu) $\downarrow$ & 5.40 & 20.43 & 19.70 & 13.28 & 100 & 2.77 & 2.33 & 1.89\\
    \bottomrule
    \end{tabular}
    \end{adjustbox}
\end{table*}

\paragraph{Hallucination with LLM} One phenomenon we observe is that zero-shot prompting with LLMs (GPT-4/GPT-3.5) to call APIs results in dire hallucination errors. These errors, while diverse, commonly manifest in erroneous behavior such as the model invoking the "AutoModel.from\_pretrained(dir\_name)" command with arbitrary GitHub repository names. Surprisingly, we also found that in TorchHub, HuggingFace and TensorFlow Hub, GPT-3.5 has less hallucination errors than GPT-4. This finding is also consistent for the settings when various retrieving methods are provided: 0-shot, BM25, GPT-Index and the oracle. This might suggest that RLHF plays a central role in turning the model to be truthful. Additional examples and discussion are in Appendix.

\subsection{Test-Time Documentation Change}

\begin{figure}[t]
    \includegraphics[width=\linewidth]{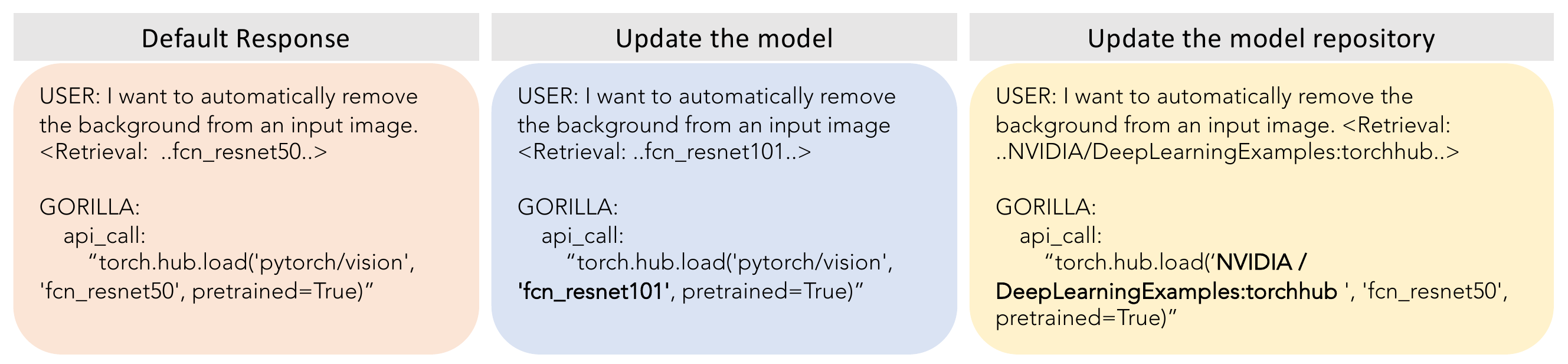}
\caption{\footnotesize \textbf{Gorilla's retriever\---aware training enables it to react to changes in the APIs.} The second column demonstrates changes in model \- upgrading FCN's ResNet\---50 backbone to ResNet\---101. The third column demonstrate changes in model registry from \texttt{pytorch/vision} to \texttt{NVIDIA/DeepLearningExamples:torchhub} }
\label{fig:docu}
\end{figure}

The rapidly evolving nature of API documentation presents a significant challenge for the application of LLMs in this field. These documents are often updated at a frequency that outpaces the re-training or fine-tuning schedule of LLMs, making these models particularly brittle to changes in the information they are designed to process. This mismatch in update frequency can lead to a decline in the utility and reliability of LLMs over time.

However, with the introduction of Gorilla's retriever-aware training, we can readily adapt to changes in API documentation. This novel approach allows the model to remain updated and relevant, even as the API documentation it relies on undergoes modifications. This is a pivotal advancement in the field, as it ensures that the LLM maintains its efficacy and accuracy over time, providing reliable outputs irrespective of changes in the underlying documentation.

For instance, consider the scenario illustrated in Figure 6, where the training of Gorilla has allowed it to react effectively to changes in APIs. This includes alterations such as upgrading the FCN's ResNet-50 backbone to ResNet-101, as demonstrated in the second column of the figure. This capability ensures that the LLM remains relevant and accurate even as the underlying models and systems undergo upgrades and improvements.
Furthermore, the third column in Figure 6 shows how Gorilla adapts to changes in the model registry from \texttt{pytorch/vision} to \texttt{NVIDIA/DeepLearningExamples:torchhub}. This reflects the model's ability to adjust to shifts in API sources, which is vital as organizations may change their preferred model registries over time.

In summary, Gorilla's ability to adapt to test-time changes in API documentation offers numerous benefits. It maintains its accuracy and relevance over time, adapts to the rapid pace of updates in API documentation, and adjusts to modifications in underlying models and systems. This makes it a robust and reliable tool for API calls, significantly enhancing its practical utility.

\subsection{API Call with Constraints}
We now focus on the language model's capability of understanding constraints. For any given task, which API call to invoke is typically a tradeoff between a multitude of factors. In the case of RESTFul APIs, it could be the cost of each invocation (\$), and the latency of response (ms), among others. Similarly, within the scope of ML APIs, it is desirable for \gorilla{} to respect constraints such as accuracy, number of learnable parameters in the model, the size on disk, peak memory consumption, FLOPS, etc. We present the underlying ablation study evaluating the ability of different models in zero-shot and with retrievers settings to respect a given accuracy constraint. This setting is best understood with an example. If the user were to ask for an Image classification model that achieves at least 80\% top-1 accuracy on the Imagenet dataset, then while both are classification models hosted by Torch Hub,    \texttt{ResNeXt-101 32x16d} with a top-1 accuracy of 84.2\% would be the right model whose API to call and not, say, \texttt{MobileNetV2} which has a top-1 accuracy of 71.88\%. 

\begin{table*}[h]
    \caption{\small \textbf{Evaluating LLMs on constraint-aware API invocations}
    }
    \label{tab:const}
    \begin{adjustbox}{max width=\textwidth}
    \begin{tabular}{l cccc|cccc|cccccc}
    \toprule
    & \multicolumn{4}{c}{GPT-3.5} & \multicolumn{4}{c}{GPT-4} & \multicolumn{4}{c}{\oursmethod{}}\\ 
     % \cmidrule(lr){4-23}
     \midrule
    & 0-shot & BM25 & GPT-Index & Oracle & 0-shot & BM25 & GPT-Index & Oracle & 0-shot & BM25 & GPT-Index & Oracle \\
    \midrule
    Torch Hub (overall) &  \textbf{73.94} & 62.67 & 81.69 & 80.98 & 62.67 & 56.33 & 71.11 & 69.01  & 71.83 & 57.04 & 71.83 & 78.16 \\
    Torch Hub (Hallu) &  19.01 & 30.98 & 14.78 & 14.08 & \textbf{15.49} & 27.46 & \textbf{14.08} & \textbf{9.15} & 19.71 & 39.43 & 26.05 & 16.90 \\
    Torch Hub (err) &  7.04 & 6.33 & 3.52 & 4.92 & 21.83 & 16.19 & 14.78 & 21.83 & 8.45 & 3.52 & 2.11 & 4.92 \\
    \midrule
    Accuracy const & 43.66 & \textbf{33.80} & \textbf{33.09} & 69.01 & 43.66 & 29.57 & 29.57 & 59.15 & \textbf{47.88} & 30.28 & 26.76 & 67.60\\
    \midrule
    & \multicolumn{4}{c}{LLAMA} & \multicolumn{4}{c}{Claude}\\ 
     % \cmidrule(lr){4-23}
     \midrule
      % \cmidrule{1-9}
    & 0-shot & BM25 & GPT-Index & Oracle & 0-shot & BM25 & GPT-Index & Oracle  \\
    % \midrule
    \cmidrule{1-9}
    Torch Hub (overall) &  0 & 8.45 & 11.97 & 19.71 & 29.92 & \textbf{81.69} & \textbf{82.39} & \textbf{81.69}\\
    Torch Hub (Hallu) &   100 & 91.54 & 88.02 & 78.87 & 67.25 & \textbf{16.19} & 15.49 & 13.38\\
    Torch Hub (err) &  0 & 0 & 0 & 1.4  & 2.81 & 2.11 & 2.11 & 4.92\\
    % \midrule
     \cmidrule{1-9}
    Accuracy const & 0 & 6.33 & 3.52 & 17.60 & 17.25 & 29.57 & 31.69 & \textbf{69.71}\\
    \bottomrule
    \end{tabular}
    \end{adjustbox}
\end{table*}

For Table~\ref{tab:const}, we filtered a subset of the Torch Hub dataset that had accuracy defined for at least one-dataset in its model card (65.26\% of TorchHub dataset in Table~\ref{tab:llm_eval}). We notice that with constraints, understandably, the accuracy drops across all models, with and without a retriever. \gorilla{} is able to match performance with the best-performing model GPT-3.5 when using retrievals (BM25, GPT-Index) and has the highest accuracy in the Zero-shot case. This highlights Gorilla's ability to navigate APIs while considering the trade-offs between different constraints.

\section{Conclusion}
\label{sec:conclusion}
LLMs are swiftly gaining popularity across diverse domains. In our study, we spotlight techniques designed to enhance the LLM's ability to accurately identify the appropriate API for a specific task—a significant but often overlooked aspect in the advancement of this technology. Since APIs function as a universal language enabling diverse systems to communicate effectively, their correct usage can boost the ability of LLMs to interact with tools in the wider world.
In this paper, we propose \oursmethod{}, a new novel pipeline for finetuning LLMs to call APIs. The finetuned model's performance surpasses prompting the state-of-the-art LLM (GPT-4) in three massive datasets we collected. \gorilla{} generates reliable API calls to ML models without hallucination, demonstrates an impressive capability to adapt to test-time API usage changes, and can satisfy constraints while picking APIs. 

\section{Limitations \& Social Impacts}
\label{sec:limitation}

With the goal of wanting to have a challenging dataset, we chose ML APIs, given their functional similarity. The potential downside to APIs that focus on the ML domain, is their propensity to produce biased predictions if trained on skewed data, potentially disadvantaging certain sub-groups. To counter this concern and foster a deeper understanding of these APIs, we are releasing our extensive dataset, consisting of over 11,000 instruction-API pairs. This resource will serve the wider community as a valuable tool for studying and benchmarking existing APIs, contributing to a more fair and optimized usage of machine learning.
\section{Acknowledgement}
\label{sec:ack}
This research is supported in part by gifts to UC Berkley Sky Computing Lab from Astronomer, Google, IBM, Intel, Lacework, Microsoft, Nexla, Samsung SDS, Uber, and VMware.

\bibliographystyle{apalike}
\bibliography{references}

\begin{thebibliography}{}

\bibitem[Ahn et~al., 2022]{ahn2022can}
Ahn, M., Brohan, A., Brown, N., Chebotar, Y., Cortes, O., David, B., Finn, C.,
  Gopalakrishnan, K., Hausman, K., Herzog, A., et~al. (2022).
\newblock Do as i can, not as i say: Grounding language in robotic affordances.
\newblock {\em arXiv preprint arXiv:2204.01691}.

\bibitem[Andor et~al., 2019]{andor2019giving}
Andor, D., He, L., Lee, K., and Pitler, E. (2019).
\newblock Giving bert a calculator: Finding operations and arguments with
  reading comprehension.
\newblock {\em arXiv preprint arXiv:1909.00109}.

\bibitem[Anthropic, 2022]{Claude}
Anthropic, h.-c. (2022).
\newblock Claude.

\bibitem[Bavishi et~al., 2019]{autopandas}
Bavishi, R., Lemieux, C., Fox, R., Sen, K., and Stoica, I. (2019).
\newblock Autopandas: neural-backed generators for program synthesis.
\newblock {\em Proceedings of the ACM on Programming Languages},
  3(OOPSLA):1--27.

\bibitem[Brown et~al., 2020]{brown2020language}
Brown, T., Mann, B., Ryder, N., Subbiah, M., Kaplan, J.~D., Dhariwal, P.,
  Neelakantan, A., Shyam, P., Sastry, G., Askell, A., et~al. (2020).
\newblock Language models are few-shot learners.
\newblock {\em Advances in neural information processing systems},
  33:1877--1901.

\bibitem[Bubeck et~al., 2023]{bubeck2023sparks}
Bubeck, S., Chandrasekaran, V., Eldan, R., Gehrke, J., Horvitz, E., Kamar, E.,
  Lee, P., Lee, Y.~T., Li, Y., Lundberg, S., et~al. (2023).
\newblock Sparks of artificial general intelligence: Early experiments with
  gpt-4.
\newblock {\em arXiv preprint arXiv:2303.12712}.

\bibitem[Chen et~al., 2021]{chen2021evaluating}
Chen, M., Tworek, J., Jun, H., Yuan, Q., Pinto, H. P. d.~O., Kaplan, J.,
  Edwards, H., Burda, Y., Joseph, N., Brockman, G., et~al. (2021).
\newblock Evaluating large language models trained on code.
\newblock {\em arXiv preprint arXiv:2107.03374}.

\bibitem[Chen et~al., 2023]{chen2023teaching}
Chen, X., Lin, M., Sch{\"a}rli, N., and Zhou, D. (2023).
\newblock Teaching large language models to self-debug.
\newblock {\em arXiv preprint arXiv:2304.05128}.

\bibitem[Chiang et~al., 2023]{vicuna}
Chiang, W.-L., Li, Z., Lin, Z., Sheng, Y., Wu, Z., Zhang, H., Zheng, L.,
  Zhuang, S., Zhuang, Y., Gonzalez, J.~E., Stoica, I., and Xing, E.~P. (2023).
\newblock Vicuna: An open-source chatbot impressing gpt-4 with 90\%* chatgpt
  quality.

\bibitem[Chowdhery et~al., 2022]{chowdhery2022palm}
Chowdhery, A., Narang, S., Devlin, J., Bosma, M., Mishra, G., Roberts, A.,
  Barham, P., Chung, H.~W., Sutton, C., Gehrmann, S., et~al. (2022).
\newblock Palm: Scaling language modeling with pathways.
\newblock {\em arXiv preprint arXiv:2204.02311}.

\bibitem[Chung et~al., 2022]{chung2022scaling}
Chung, H.~W., Hou, L., Longpre, S., Zoph, B., Tay, Y., Fedus, W., Li, E., Wang,
  X., Dehghani, M., Brahma, S., et~al. (2022).
\newblock Scaling instruction-finetuned language models.
\newblock {\em arXiv preprint arXiv:2210.11416}.

\bibitem[Cobbe et~al., 2021]{cobbe2021training}
Cobbe, K., Kosaraju, V., Bavarian, M., Chen, M., Jun, H., Kaiser, L., Plappert,
  M., Tworek, J., Hilton, J., Nakano, R., et~al. (2021).
\newblock Training verifiers to solve math word problems.
\newblock {\em arXiv preprint arXiv:2110.14168}.

\bibitem[Devlin et~al., 2017]{devlin2017robustfill}
Devlin, J., Uesato, J., Bhupatiraju, S., Singh, R., Mohamed, A.-r., and Kohli,
  P. (2017).
\newblock Robustfill: Neural program learning under noisy i/o.
\newblock In {\em International conference on machine learning}, pages
  990--998. PMLR.

\bibitem[Gao et~al., 2022]{gao2022pal}
Gao, L., Madaan, A., Zhou, S., Alon, U., Liu, P., Yang, Y., Callan, J., and
  Neubig, G. (2022).
\newblock Pal: Program-aided language models.
\newblock {\em arXiv preprint arXiv:2211.10435}.

\bibitem[Iyer et~al., 2022]{iyer2022opt}
Iyer, S., Lin, X.~V., Pasunuru, R., Mihaylov, T., Simig, D., Yu, P., Shuster,
  K., Wang, T., Liu, Q., Koura, P.~S., et~al. (2022).
\newblock Opt-iml: Scaling language model instruction meta learning through the
  lens of generalization.
\newblock {\em arXiv preprint arXiv:2212.12017}.

\bibitem[Jain et~al., 2022]{jain2022jigsaw}
Jain, N., Vaidyanath, S., Iyer, A., Natarajan, N., Parthasarathy, S., Rajamani,
  S., and Sharma, R. (2022).
\newblock Jigsaw: Large language models meet program synthesis.
\newblock In {\em Proceedings of the 44th International Conference on Software
  Engineering}, pages 1219--1231.

\bibitem[Kim et~al., 2023]{kim2023language}
Kim, G., Baldi, P., and McAleer, S. (2023).
\newblock Language models can solve computer tasks.
\newblock {\em arXiv preprint arXiv:2303.17491}.

\bibitem[Kojima et~al., 2022]{kojima2022large}
Kojima, T., Gu, S.~S., Reid, M., Matsuo, Y., and Iwasawa, Y. (2022).
\newblock Large language models are zero-shot reasoners.
\newblock {\em arXiv preprint arXiv:2205.11916}.

\bibitem[Komeili et~al., 2021]{komeili2021internet}
Komeili, M., Shuster, K., and Weston, J. (2021).
\newblock Internet-augmented dialogue generation.
\newblock {\em arXiv preprint arXiv:2107.07566}.

\bibitem[Lachaux et~al., 2020]{lachaux2020unsupervised}
Lachaux, M.-A., Roziere, B., Chanussot, L., and Lample, G. (2020).
\newblock Unsupervised translation of programming languages.
\newblock {\em arXiv preprint arXiv:2006.03511}.

\bibitem[Lazaridou et~al., 2022]{lazaridou2022internet}
Lazaridou, A., Gribovskaya, E., Stokowiec, W., and Grigorev, N. (2022).
\newblock Internet-augmented language models through few-shot prompting for
  open-domain question answering.
\newblock {\em arXiv preprint arXiv:2203.05115}.

\bibitem[Li et~al., 2023]{li2023starcoder}
Li, R., Allal, L.~B., Zi, Y., Muennighoff, N., Kocetkov, D., Mou, C., Marone,
  M., Akiki, C., Li, J., Chim, J., et~al. (2023).
\newblock Starcoder: may the source be with you!
\newblock {\em arXiv preprint arXiv:2305.06161}.

\bibitem[Li et~al., 2022]{li2022competition}
Li, Y., Choi, D., Chung, J., Kushman, N., Schrittwieser, J., Leblond, R.,
  Eccles, T., Keeling, J., Gimeno, F., Dal~Lago, A., et~al. (2022).
\newblock Competition-level code generation with alphacode.
\newblock {\em Science}, 378(6624):1092--1097.

\bibitem[Liang et~al., 2023]{liang2023taskmatrix}
Liang, Y., Wu, C., Song, T., Wu, W., Xia, Y., Liu, Y., Ou, Y., Lu, S., Ji, L.,
  Mao, S., et~al. (2023).
\newblock Taskmatrix. ai: Completing tasks by connecting foundation models with
  millions of apis.
\newblock {\em arXiv preprint arXiv:2303.16434}.

\bibitem[Menon et~al., 2013]{flashfill}
Menon, A., Tamuz, O., Gulwani, S., Lampson, B., and Kalai, A. (2013).
\newblock A machine learning framework for programming by example.
\newblock In {\em International Conference on Machine Learning}, pages
  187--195. PMLR.

\bibitem[Nakano et~al., 2021]{nakano2021webgpt}
Nakano, R., Hilton, J., Balaji, S., Wu, J., Ouyang, L., Kim, C., Hesse, C.,
  Jain, S., Kosaraju, V., Saunders, W., et~al. (2021).
\newblock Webgpt: Browser-assisted question-answering with human feedback.
\newblock {\em arXiv preprint arXiv:2112.09332}.

\bibitem[Nijkamp et~al., 2023]{nijkamp2023codegen2}
Nijkamp, E., Hayashi, H., Xiong, C., Savarese, S., and Zhou, Y. (2023).
\newblock Codegen2: Lessons for training llms on programming and natural
  languages.
\newblock {\em arXiv preprint arXiv:2305.02309}.

\bibitem[Nijkamp et~al., 2022]{nijkamp2022codegen}
Nijkamp, E., Pang, B., Hayashi, H., Tu, L., Wang, H., Zhou, Y., Savarese, S.,
  and Xiong, C. (2022).
\newblock Codegen: An open large language model for code with multi-turn
  program synthesis.
\newblock {\em arXiv preprint arXiv:2203.13474}.

\bibitem[OpenAI, 2023]{openai2023gpt4}
OpenAI (2023).
\newblock Gpt-4 technical report.

\bibitem[OpenAI and https://openai.com/blog/chatgpt, 2022]{ChatGPT}
OpenAI and https://openai.com/blog/chatgpt (2022).
\newblock Chatgpt.

\bibitem[Sanh et~al., 2021]{sanh2021multitask}
Sanh, V., Webson, A., Raffel, C., Bach, S.~H., Sutawika, L., Alyafeai, Z.,
  Chaffin, A., Stiegler, A., Scao, T.~L., Raja, A., et~al. (2021).
\newblock Multitask prompted training enables zero-shot task generalization.
\newblock {\em arXiv preprint arXiv:2110.08207}.

\bibitem[Scao et~al., 2022]{scao2022bloom}
Scao, T.~L., Fan, A., Akiki, C., Pavlick, E., Ili{\'c}, S., Hesslow, D.,
  Castagn{\'e}, R., Luccioni, A.~S., Yvon, F., Gall{\'e}, M., et~al. (2022).
\newblock Bloom: A 176b-parameter open-access multilingual language model.
\newblock {\em arXiv preprint arXiv:2211.05100}.

\bibitem[Schick et~al., 2023]{schick2023toolformer}
Schick, T., Dwivedi-Yu, J., Dess{\`\i}, R., Raileanu, R., Lomeli, M.,
  Zettlemoyer, L., Cancedda, N., and Scialom, T. (2023).
\newblock Toolformer: Language models can teach themselves to use tools.
\newblock {\em arXiv preprint arXiv:2302.04761}.

\bibitem[Schick and Sch{\"u}tze, 2020]{schick2020exploiting}
Schick, T. and Sch{\"u}tze, H. (2020).
\newblock Exploiting cloze questions for few shot text classification and
  natural language inference.
\newblock {\em arXiv preprint arXiv:2001.07676}.

\bibitem[Shen et~al., 2023]{shen2023hugginggpt}
Shen, Y., Song, K., Tan, X., Li, D., Lu, W., and Zhuang, Y. (2023).
\newblock Hugginggpt: Solving ai tasks with chatgpt and its friends in
  huggingface.
\newblock {\em arXiv preprint arXiv:2303.17580}.

\bibitem[Shinn et~al., 2023]{shinn2023reflexion}
Shinn, N., Labash, B., and Gopinath, A. (2023).
\newblock Reflexion: an autonomous agent with dynamic memory and
  self-reflection.
\newblock {\em arXiv preprint arXiv:2303.11366}.

\bibitem[Shuster et~al., 2022]{shuster2022blenderbot}
Shuster, K., Xu, J., Komeili, M., Ju, D., Smith, E.~M., Roller, S., Ung, M.,
  Chen, M., Arora, K., Lane, J., et~al. (2022).
\newblock Blenderbot 3: a deployed conversational agent that continually learns
  to responsibly engage.
\newblock {\em arXiv preprint arXiv:2208.03188}.

\bibitem[Taori et~al., 2023]{alpaca}
Taori, R., Gulrajani, I., Zhang, T., Dubois, Y., Li, X., Guestrin, C., Liang,
  P., and Hashimoto, T.~B. (2023).
\newblock Stanford alpaca: An instruction-following llama model.
\newblock \url{https://github.com/tatsu-lab/stanford_alpaca}.

\bibitem[Thoppilan et~al., 2022]{thoppilan2022lamda}
Thoppilan, R., De~Freitas, D., Hall, J., Shazeer, N., Kulshreshtha, A., Cheng,
  H.-T., Jin, A., Bos, T., Baker, L., Du, Y., et~al. (2022).
\newblock Lamda: Language models for dialog applications.
\newblock {\em arXiv preprint arXiv:2201.08239}.

\bibitem[Touvron et~al., 2023]{touvron2023llama}
Touvron, H., Lavril, T., Izacard, G., Martinet, X., Lachaux, M.-A., Lacroix,
  T., Rozi{\`e}re, B., Goyal, N., Hambro, E., Azhar, F., et~al. (2023).
\newblock Llama: Open and efficient foundation language models.
\newblock {\em arXiv preprint arXiv:2302.13971}.

\bibitem[Vemprala et~al., 2023]{vemprala2023chatgpt}
Vemprala, S., Bonatti, R., Bucker, A., and Kapoor, A. (2023).
\newblock Chatgpt for robotics: Design principles and model abilities.
\newblock {\em 2023}.

\bibitem[Wang et~al., 2022a]{wang2022self}
Wang, Y., Kordi, Y., Mishra, S., Liu, A., Smith, N.~A., Khashabi, D., and
  Hajishirzi, H. (2022a).
\newblock Self-instruct: Aligning language model with self generated
  instructions.
\newblock {\em arXiv preprint arXiv:2212.10560}.

\bibitem[Wang et~al., 2022b]{wang2022super}
Wang, Y., Mishra, S., Alipoormolabashi, P., Kordi, Y., Mirzaei, A., Naik, A.,
  Ashok, A., Dhanasekaran, A.~S., Arunkumar, A., Stap, D., et~al. (2022b).
\newblock Super-naturalinstructions: Generalization via declarative
  instructions on 1600+ nlp tasks.
\newblock In {\em Proceedings of the 2022 Conference on Empirical Methods in
  Natural Language Processing}, pages 5085--5109.

\bibitem[Wei et~al., 2022]{wei2022chain}
Wei, J., Wang, X., Schuurmans, D., Bosma, M., Chi, E., Le, Q., and Zhou, D.
  (2022).
\newblock Chain of thought prompting elicits reasoning in large language
  models.
\newblock {\em arXiv preprint arXiv:2201.11903}.

\bibitem[Xu et~al., 2022]{xu2022systematic}
Xu, F.~F., Alon, U., Neubig, G., and Hellendoorn, V.~J. (2022).
\newblock A systematic evaluation of large language models of code.
\newblock In {\em Proceedings of the 6th ACM SIGPLAN International Symposium on
  Machine Programming}, pages 1--10.

\bibitem[Yao et~al., 2022]{yao2022react}
Yao, S., Zhao, J., Yu, D., Du, N., Shafran, I., Narasimhan, K., and Cao, Y.
  (2022).
\newblock React: Synergizing reasoning and acting in language models.
\newblock {\em arXiv preprint arXiv:2210.03629}.

\bibitem[Zeng et~al., 2022]{zeng2022glm}
Zeng, A., Liu, X., Du, Z., Wang, Z., Lai, H., Ding, M., Yang, Z., Xu, Y.,
  Zheng, W., Xia, X., et~al. (2022).
\newblock Glm-130b: An open bilingual pre-trained model.
\newblock {\em arXiv preprint arXiv:2210.02414}.

\bibitem[Zhang et~al., 2022]{zhang2022opt}
Zhang, S., Roller, S., Goyal, N., Artetxe, M., Chen, M., Chen, S., Dewan, C.,
  Diab, M., Li, X., Lin, X.~V., et~al. (2022).
\newblock Opt: Open pre-trained transformer language models.
\newblock {\em arXiv preprint arXiv:2205.01068}.

\end{thebibliography}

\newpage
\newpage
\section{Appendix}

\subsection{Dataset Details}
Our dataset is multi-faceted, comprising three distinct domains: Torch Hub, Tensor Hub, and HuggingFace. Each entry within this dataset is rich in detail, carrying critical pieces of information that further illuminate the nature of the data. Delving deeper into the specifics of each domain, Torch Hub provides 95 APIs. The second domain, Tensor Hub, is more expansive with a total of 696 APIs. Finally, the most extensive of them all, HuggingFace, comprises 925 APIs.

To enhance the value and utility of our dataset, we've undertaken an additional initiative. With each API, we have generated a set of 10 unique instructions. These instructions, carefully crafted and meticulously tailored, serve as a guide for both training and evaluation. This initiative ensures that every API is not just represented in our dataset, but is also comprehensively understood and effectively utilizable.

In essence, our dataset is more than just a collection of APIs across three domains. It is a comprehensive resource, carefully structured and enriched with added layers of guidance and evaluation parameters.

\paragraph{Domain Classification} The unique domain names encompassed within our dataset are illustrated in Figure~\ref{fig:domains}. The dataset consists of three sources with a diverse range of domains: Torch Hub houses 6 domains, Tensor Hub accommodates a much broader selection with 57 domains, while HuggingFace incorporates 37 domains. To exemplify the structure and nature of our dataset, we invite you to refer to the domain names represented in Figure~\ref{fig:data_exp}.

\paragraph{API Call Task} In this task, we test the model's capability to generate a single line of code, either in a zero-shot fashion or by leveraging an API reference. Primarily designed for evaluation purposes, this task effectively gauges the model's proficiency in identifying and utilizing the appropriate API call. 

\paragraph{API Provider Component} This facet relates to the provision of the programming language. In this context, the API provider plays a vital role as it serves as a foundation upon which APIs are built and executed.

\paragraph{Explanation Element} This component offers valuable insights into the rationale behind the usage of a particular API, detailing how it aligns with the prescribed requirements. Furthermore, when certain constraints are imposed, this segment also incorporates those limitations. Thus, the explanation element serves a dual purpose, offering a deep understanding of API selection, as well as the constraints that might influence such a selection. This balanced approach ensures a comprehensive understanding of the API usage within the given context.

\paragraph{Code} Example code for accomplishing the task. We de-prioritize this as we haven't tested the execution result of the code. We leave this for future works, but make this data available in-case others want to build on it.

\begin{figure}
\begin{tcolorbox}
    \textbf{Torch Hub domain names}: Classification, Semantic Segmentation, Object Detection, Audio Separation, Video Classification, Text-to-Speech
\end{tcolorbox}
\begin{tcolorbox}
    \textbf{Tensor Hub domain names}: text-sequence-alignment, text-embedding, text-language-model, text-preprocessing, text-classification, text-generation, text-question-answering, text-retrieval-question-answering, text-segmentation, text-to-mel, image-classification, image-feature-vector, image-object-detection, image-segmentation, image-generator, image-pose-detection, image-rnn-agent, image-augmentation, image-classifier, image-style-transfer, image-aesthetic-quality, image-depth-estimation, image-super-resolution, image-deblurring, image-extrapolation, image-text-recognition, image-dehazing, image-deraining, image-enhancemenmt, image-classification-logits, image-frame-interpolation, image-text-detection, image-denoising, image-others, video-classification, video-feature-extraction, video-generation, video-audio-text, video-text, audio-embedding, audio-event-classification, audio-command-detection, audio-paralinguists-classification, audio-speech-to-text, audio-speech-synthesis, audio-synthesis, audio-pitch-extraction
\end{tcolorbox}
\begin{tcolorbox}
    \textbf{HuggingFace domain names}: Multimodal Feature Extraction, Multimodal Text-to-Image, Multimodal Image-to-Text, Multimodal Text-to-Video, Multimodal Visual Question Answering, Multimodal Document Question Answer, Multimodal Graph Machine Learning, Computer Vision Depth Estimation, Computer Vision Image Classification, Computer Vision Object Detection, Computer Vision Image Segmentation, Computer Vision Image-to-Image, Computer Vision Unconditional Image Generation, Computer Vision Video Classification, Computer Vision Zero-Shor Image Classification, Natural Language Processing Text Classification, Natural Language Processing Token Classification, Natural Language Processing Table Question Answering, Natural Language Processing Question Answering, Natural Language Processing Zero-Shot Classification, Natural Language Processing Translation, Natural Language Processing Summarization, Natural Language Processing Conversational, Natural Language Processing Text Generation, Natural Language Processing Fill-Mask, Natural Language Processing Text2Text Generation, Natural Language Processing Sentence Similarity, Audio Text-to-Speech, Audio Automatic Speech Recognition, Audio Audio-to-Audio, Audio Audio Classification, Audio Voice Activity Detection, Tabular Tabular Classification, Tabular Tabular Regression, Reinforcement Learning Reinforcement Learning, Reinforcement Learning Robotics
\end{tcolorbox}
\caption{\textbf{Domain names}: Domain names with the three dataset. Tensor Hub is the smallest dataset while the other two hubs contain many more models.}
    \label{fig:domains}
\end{figure}

\begin{figure}
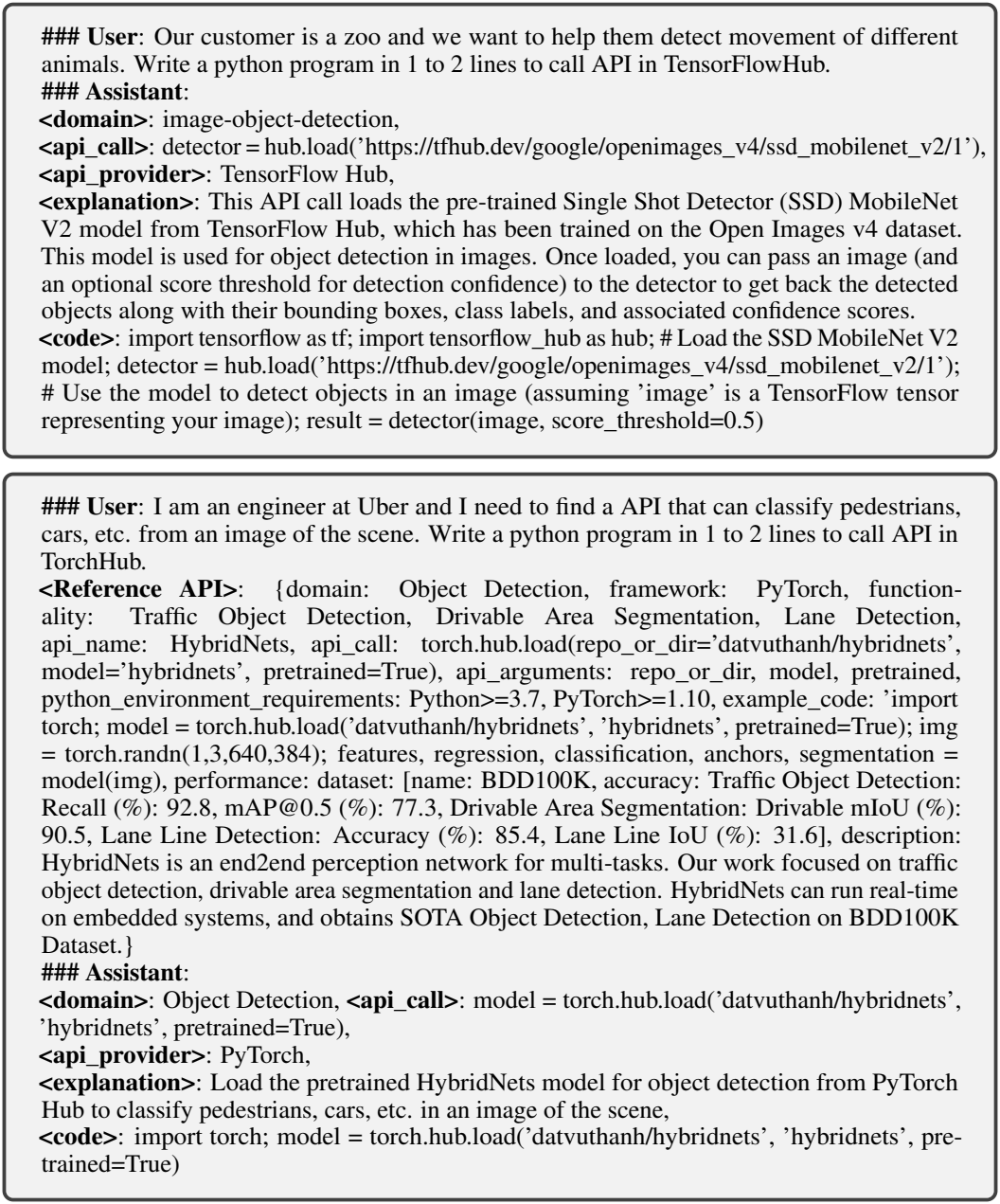

\begin{tcolorbox}
    \textbf{\#\#\# User}: Our customer is a zoo and we want to help them detect movement of different animals. Write a python program in 1 to 2 lines to call API in TensorFlowHub.
    
    \textbf{\#\#\# Assistant}: 
    
    \textbf{<domain>}: image-object-detection,
    
    \textbf{<api\_call>}: detector = hub.load('https://tfhub.dev/google/openimages\_v4/ssd\_mobilenet\_v2/1'),
    \textbf{<api\_provider>}: TensorFlow Hub,
    
    \textbf{<explanation>}: This API call loads the pre-trained Single Shot Detector (SSD) MobileNet V2 model from TensorFlow Hub, which has been trained on the Open Images v4 dataset. This model is used for object detection in images. Once loaded, you can pass an image (and an optional score threshold for detection confidence) to the detector to get back the detected objects along with their bounding boxes, class labels, and associated confidence scores.
    
    \textbf{<code>}: import tensorflow as tf; import tensorflow\_hub as hub; \# Load the SSD MobileNet V2 model; detector = hub.load('https://tfhub.dev/google/openimages\_v4/ssd\_mobilenet\_v2/1'); \# Use the model to detect objects in an image (assuming 'image' is a TensorFlow tensor representing your image); result = detector(image, score\_threshold=0.5)
\end{tcolorbox}
\begin{tcolorbox}
    \textbf{\#\#\# User}: I am an engineer at Uber and I need to find a API that can classify pedestrians, cars, etc. from an image of the scene. Write a python program in 1 to 2 lines to call API in TorchHub.

    \textbf{<Reference API>}: \{domain: Object Detection, framework: PyTorch, functionality: Traffic Object Detection, Drivable Area Segmentation, Lane Detection, api\_name: HybridNets, api\_call: torch.hub.load(repo\_or\_dir='datvuthanh/hybridnets', model='hybridnets', pretrained=True), api\_arguments: repo\_or\_dir, model, pretrained, python\_environment\_requirements: Python>=3.7, PyTorch>=1.10, example\_code: 'import torch; model = torch.hub.load('datvuthanh/hybridnets', 'hybridnets', pretrained=True); img = torch.randn(1,3,640,384); features, regression, classification, anchors, segmentation = model(img), performance: {dataset: [{name: BDD100K, accuracy: {Traffic Object Detection: {Recall (\%): 92.8, mAP@0.5 (\%): 77.3}, Drivable Area Segmentation: {Drivable mIoU (\%): 90.5}, Lane Line Detection: {Accuracy (\%): 85.4, Lane Line IoU (\%): 31.6}}}]}, description: HybridNets is an end2end perception network for multi-tasks. Our work focused on traffic object detection, drivable area segmentation and lane detection. HybridNets can run real-time on embedded systems, and obtains SOTA Object Detection, Lane Detection on BDD100K Dataset.\}

    \textbf{\#\#\# Assistant}: 
    
    \textbf{<domain>}: Object Detection, \textbf{<api\_call>}: model = torch.hub.load('datvuthanh/hybridnets', 'hybridnets', pretrained=True),
    
    \textbf{<api\_provider>}: PyTorch,
    
    \textbf{<explanation>}: Load the pretrained HybridNets model for object detection from PyTorch Hub to classify pedestrians, cars, etc. in an image of the scene, 
    
    \textbf{<code>}: import torch; model = torch.hub.load('datvuthanh/hybridnets', 'hybridnets', pretrained=True)
\end{tcolorbox}
    \caption{\textbf{Example of the Dataset}: Two examples of the dataset, the above one is zero-shot (without information retrievers) and the bottom one is with information retriever. }
    \label{fig:data_exp}
\end{figure}

\subsection{\gorilla{} Details}
We provide all the training details for \gorilla{} in this section. This includes how we divide up the training, evaluation dataset, training hyperparameters for \gorilla{}.

\paragraph{Data} For HuggingFace, we devise the entire dataset into 90\% training and 10\% evaluation. For Torch Hub and Tensor Hub, we devise the data in to 80\% training and 20\% testing. 

\paragraph{Training}
We train \gorilla for 5 epochs with the 2e-5 learning rate with cosine decay. The details are provide in Tab.~\ref{tab:hyperparameter}. We finetune it on 8xA100 with 40G memory each. 

\begin{table*}[!htb]
% \color{red}
\caption{Hyperparameters for training \gorilla{} }
\footnotesize
\setlength\tabcolsep{3.5pt}
\label{tab:hyperparameter}
\centering
\begin{tabular}{p{5cm}p{3cm}p{3cm}p{2.5cm}p{2.5cm}p{2cm}p{2cm}}
% {lcccccccccccc}
\toprule
Hyperparameter Name & Value \\ 
\midrule
learning rate & 2e-5 \\
batch size & 64 \\
epochs & 5 \\
warmup ratio & 0.03 \\
weight decay & 0 \\
max seq length & 2048 \\

\bottomrule 
\end{tabular}
\end{table*}

\subsection{Performance Comparison}
We provide a full comparison of each model's performance in this section. In Fig~\ref{fig:full1} and Fig.~\ref{fig:full2}, the full set of comparisons is provided. We see that especially in zero-shot case, \gorilla{} surpasses the GPT-4 and GPT-3.5 by a large margin. The GPT-4 and GPT-3.5 gets around 40\% accuracy in Torch Hub and Tensor Hub, which are two structured API calls. Compared to that, HuggingFace is a more flexible and diverse Hub, as a result, the performance on HuggingFace is not as competitive.

\subsubsection{Evaluation}
For ease of evaluation, we manually cleaned up the dataset to make sure each API call domain only contains the valid call in the form of: 
\begin{tcolorbox}
\centering
    API\_name(API\_$\mathrm{arg_{1}}$, API\_$\mathrm{arg_{2}}$, ..., API\_$\mathrm{arg_{k}}$)
\end{tcolorbox}
Our framework allows the user to define any combination of the arguments to check. For Torch Hub, we check for the API name \texttt{torch.hub.load} with arguments \texttt{repo\_or\_dir} and \texttt{model}. For Tensor Hub, we check API name \texttt{hub.KerasLayer} and \texttt{hub.load} with argument \texttt{handle}. For HuggingFace, since there are many API function names, we don't list all of them here. One specific note is that we require the \texttt{pretrained\_model\_name\_or\_path} argument for all the calls except for \texttt{pipeline}. For \texttt{pipeline}, we don't require the \texttt{pretrained\_model\_name\_or\_path} argument since it automatically select a model for you once \texttt{task} is specified.  

\subsubsection{Hallucination}
We found especially in HuggingFace, the GPT-4 model incurs serious hallucination problems. It would sometimes put a GitHub name that is not associated with the HuggingFace repository in to the domain of \texttt{pretrained\_model\_name\_or\_path}. Fig.~\ref{fig:hallucination} demonstrates some examples and we also observe that GPT-4 sometimes assumes the user have a local path to the model like \texttt{your\_model\_name}. This is greatly reduced by \gorilla{} as we see the hallucination error comparison in Tab.~\ref{tab:llm_eval}.

\begin{figure}
\begin{tcolorbox}
    generate\_video = pipeline("text-to-video", model="your\_model\_name")
\end{tcolorbox}
\begin{tcolorbox}
    vqa = pipeline("visual-question-answering", model="microsoft/clip-vqa-base", tokenizer="microsoft/clip-vqa-base")
\end{tcolorbox}
\begin{tcolorbox}
    depth\_estimator = pipeline("depth-estimation", model="intel-isl/MiDaS", tokenizer="intel-isl/MiDaS")
\end{tcolorbox}
    \caption{\textbf{Hallucination Examples}: GPT-4 incurs serious hallucination errors in HuggingFace. We show a couple of examples in the figure.}
    \label{fig:hallucination}
\end{figure}

\begin{figure}[t]
    \includegraphics[width=\linewidth]{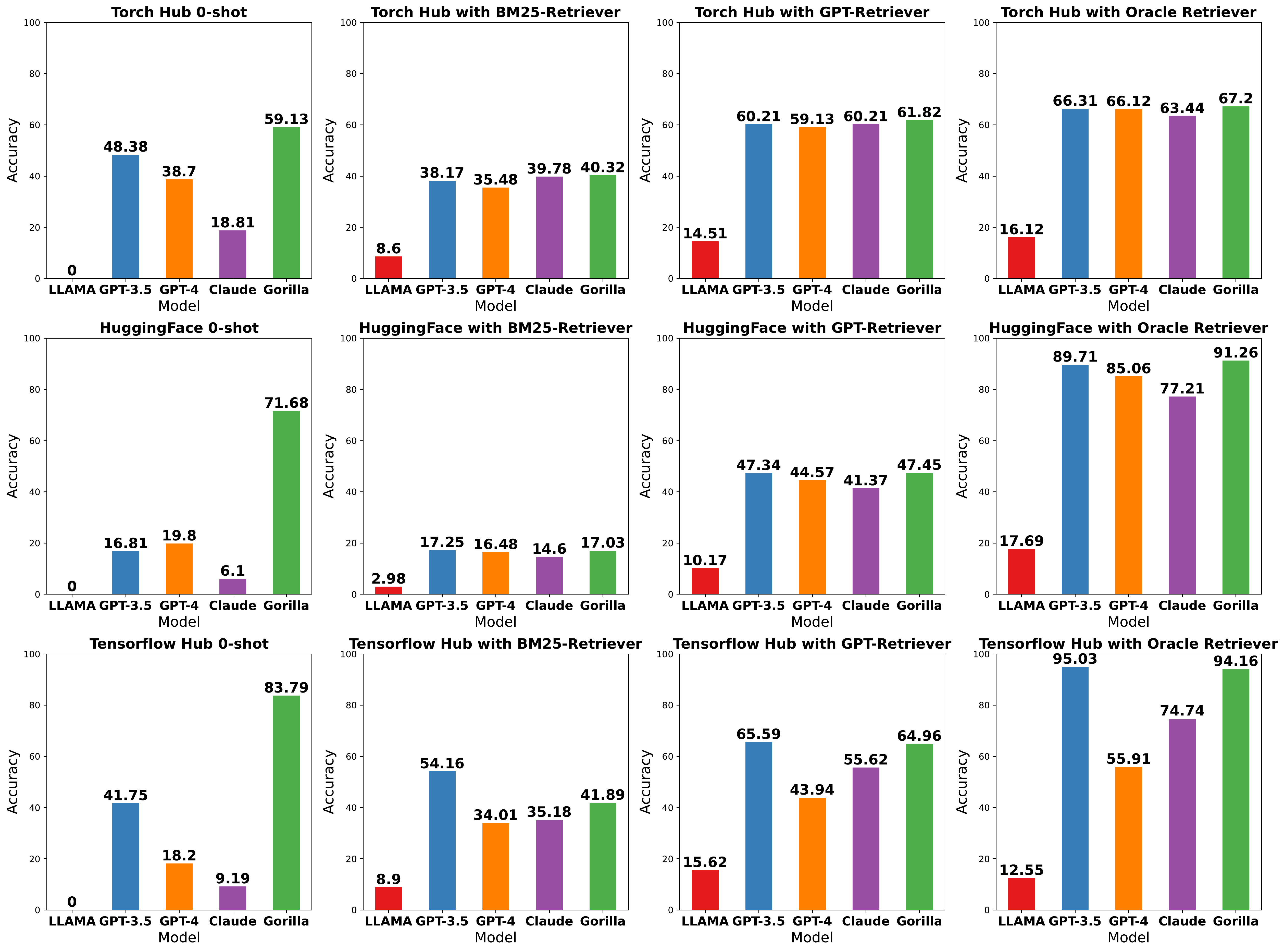}
\caption{\footnotesize \textbf{ Performance}: We plot each model's performance on different configurations. We see that \gorilla{} performs extremely well in the zero-shot setting. While even when the oracle answer is given, \gorilla{} is still the best.}
\label{fig:full1}
\end{figure}

\begin{figure}[t]
    \includegraphics[width=\linewidth]{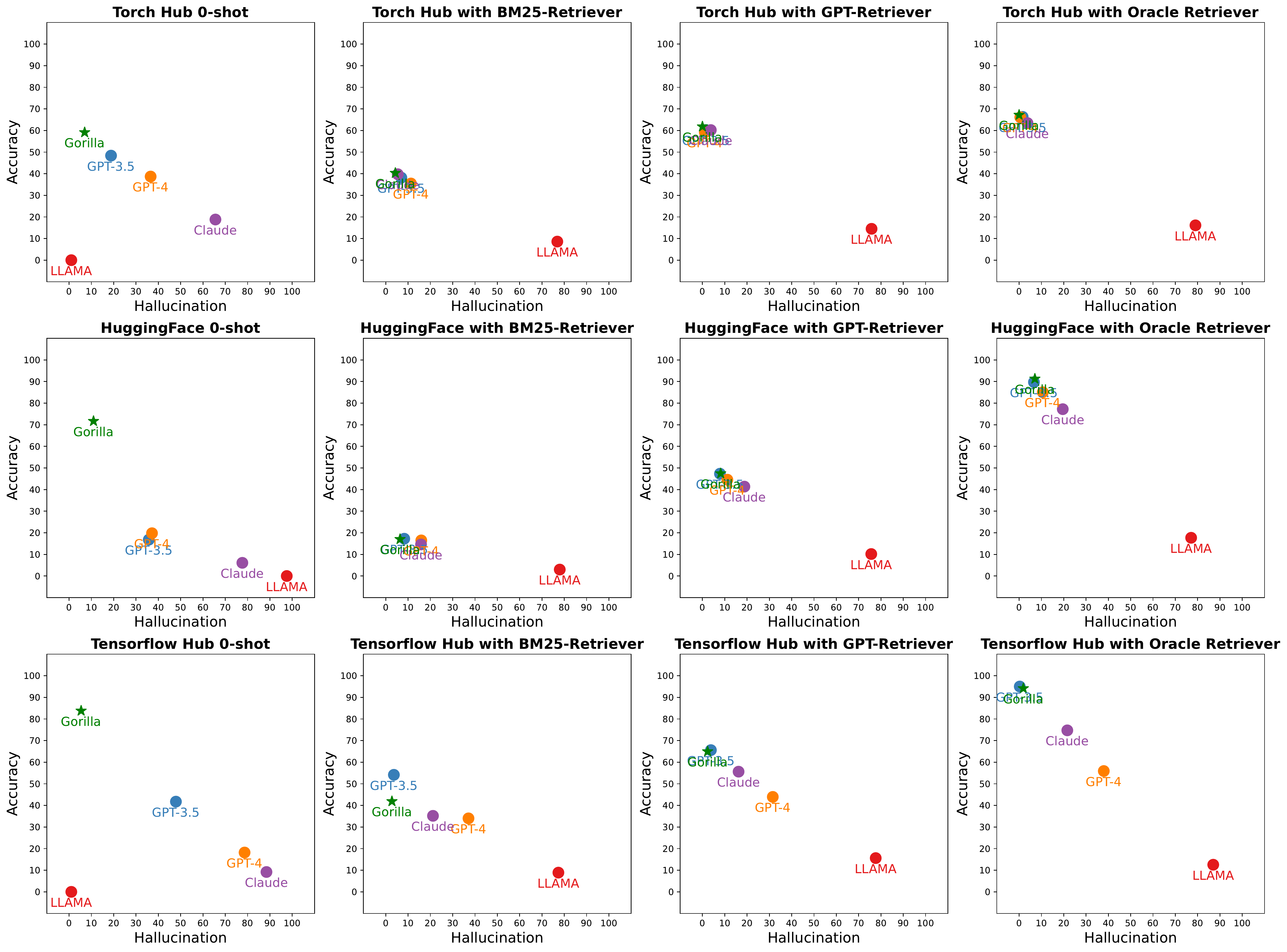}
\caption{\footnotesize \textbf{ Accuracy vs Hallucination}: We plot each model's performance on different configurations. We found that in the zero-shot setting, \gorilla{} has the most accuracy gain while maintaining good factual capability. When prompting with different retrievers, \gorilla{} is still capable to avoid the hallucination errors.}
\label{fig:full2}
\end{figure}

%%%%%%%%%%%%%%%%%%%%%%%%%%%%%%%%%%%%%%%%%%%%%%%%%%%%%%%%%%%%

\end{document}